\PassOptionsToPackage{table,svgnames,dvipsnames}{xcolor}

\documentclass[a4paper, 10pt, conference]{ieeeconf}      

\IEEEoverridecommandlockouts                              

\renewcommand{\baselinestretch}{0.948} 
\usepackage[export]{adjustbox}
\usepackage{float}

\usepackage{graphicx} 
\usepackage{mathtools}
\usepackage{amsmath} 
\usepackage{amssymb}  
\usepackage{bm}
\usepackage{mathrsfs}
\usepackage{hyperref}
\usepackage{tikz}
\usepackage{pgfplots}
\usepackage{pgfplotstable}
\usepackage{xcolor}
\usepackage{caption}
\usepackage{subcaption}
\usepackage{siunitx}
\usepackage[normalem]{ulem}
\usepackage[%
  backend=biber,
  url=true,
  style=ieee, 
  sorting=none, 
  maxnames=4,
  minnames=3,
  maxbibnames=99,
  giveninits,
  uniquename=init]{biblatex} 
\usepackage{booktabs}

\graphicspath{
    {figures/}
}

\title{\LARGE \bf
Learning Robust Agile Flight Control with Stability Guarantees
}

\author{Lukas Pries$^{1}$ and Markus Ryll$^{1}$
\thanks{$^{1}$Lukas Pries and Markus Ryll are with the Autonomous Aerial Systems Lab, Dep. of Aerospace and Geodesy,
        TU Munich, Germany,
        }%
}

\bibliography{bibliography}

\begin{document}

\maketitle
\thispagestyle{empty}
\pagestyle{empty}

\newcommand{\tikzcircle}[2][black,fill=black]{\tikz[baseline=-0.5ex]\draw[#1,radius=#2] (0,0) circle ;} 
\newcommand{\bb}[1]{\mathbf{#1}}
\newcommand{\MR}[1]{\textcolor{red}{#1}}
\newcommand{\MRN}[1]{\textcolor{red}{\sout{#1}}}

\begin{abstract}
In the evolving landscape of high-speed agile quadrotor flight, achieving precise trajectory tracking at the platform's operational limits is paramount. Controllers must handle actuator constraints, exhibit robustness to disturbances, and remain computationally efficient for safety-critical applications. In this work, we present a novel neural-augmented feedback controller for agile flight control. The controller addresses individual limitations of existing state-of-the-art control paradigms and unifies their strengths. We demonstrate the controller's capabilities, including the accurate tracking of highly aggressive trajectories that surpass the feasibility of the actuators. Notably, the controller provides universal stability guarantees, enhancing its robustness and tracking performance even in exceedingly disturbance-prone settings. Its nonlinear feedback structure is highly efficient enabling fast computation at high update rates. Moreover, the learning process in simulation is both fast and stable, and the controller's inherent robustness allows direct deployment to real-world platforms without the need for training augmentations or fine-tuning.

\end{abstract}

\section{INTRODUCTION}

In aerial robotics, quadrotors have gained prominence for their impressive agility – the ability to perform rapid, precise maneuvers in complex environments. This agility is crucial, especially in time-sensitive missions, including search and rescue operations, surveillance, exploration, drone delivery, or competitive drone racing \cite{hanover2023autonomous}.

However, agile flight presents significant challenges, including nonlinear dynamics, aerodynamic effects, unknown disturbances and physical actuator limitations. To safely execute high-speed trajectories in cluttered environments, an accurate and robust trajectory-tracking controller is essential.
Furthermore, stability and robustness guarantees are highly desirable for deploying aerial systems in safety-critical contexts, rendering agile flight a pivotal control benchmark  \cite{foehn2022agilicious,sun2022comparative,kaufmann2018deep,kaufmann2022benchmark}.

Despite extensive research, state-of-the-art control paradigms still have critical limitations:

\textit{Nonlinear feedback controllers} have fundamental stability and robustness properties and show impressive performance in tracking high-speed trajectories \cite{tal2020accurate, faessler2017differential}. However, effectively handling actuator limits remains an open challenge \cite{sun2022comparative}. Conversely, \textit{predictive methods} like Nonlinear Model Predictive Control (NMPC) excel in handling actuator limits, with its predictive capabilities being touted as advantageous for tracking high speeds trajectories \cite{bicego2020nonlinear}. This is highlighted in a recent study on tracking trajectories exceeding actuator feasibility \cite{sun2022comparative}. However, solving a nonlinear optimization problem at each iteration is computationally demanding and its non-convex nature poses reliability issues. 

A computationally efficient yet expressive alternative exists in \textit{neural policies} which can approximate optimal control inputs. However, poor stability and robustness properties limit their applicability.

\begin{figure}[t!]
    \centering
    \includegraphics[width=0.475\textwidth]{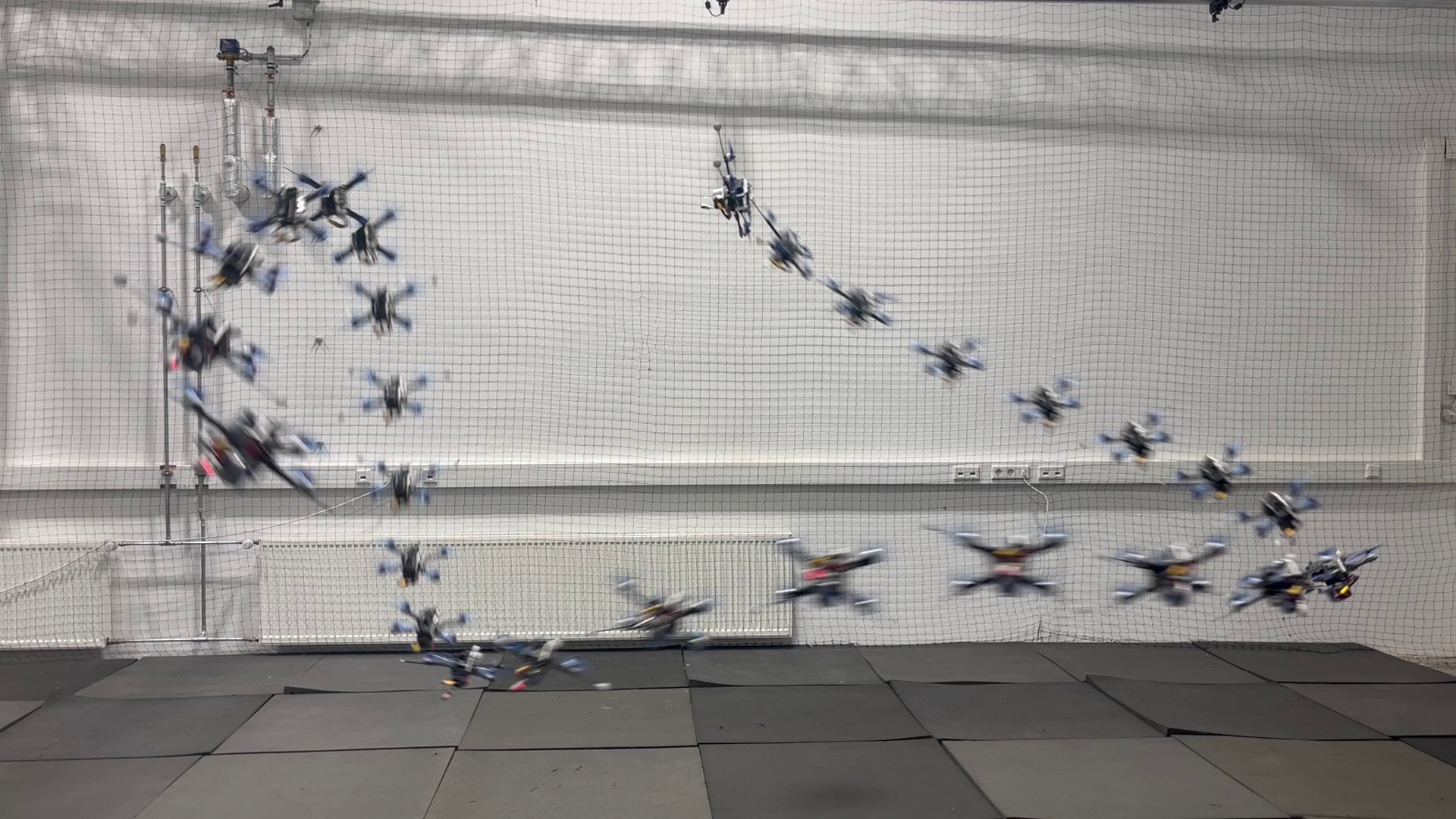}
    \caption{Experimental demonstration of a quadrotor tracking a highly dynamic racing trajectory, highlighting the controller’s accuracy and robustness in aggressive flight.}
    \label{fig:agileflight}
    \vspace{-5mm}
\end{figure}

In this work, we present a novel tracking controller which is optimal with respect to its nonlinear costs, has stability guarantees and is computationally efficient. Therefore, overcoming limitations and unifying the strengths of existing state-of-the-art control paradigms.

\subsection{Related Work}
    We compare our method to state-of-the-art control methods for agile flight including \textit{Nonlinear Feedback Control} and \textit{Model Predictive Control} and \textit{Neural Policy Control}. 
    In the following, we review works that advance trajectory tracking in all three categories.
    
    \paragraph{\textbf{Nonlinear Feedback Control}}
    Early work in quadrotor control focused on stabilizing flight in hovering and near-hovering conditions. Linear control methods such as PID and LQR perform sufficiently good near the hover state \cite{bouabdallah2004pid, castillo2005stabilization}. However, their tracking performance significantly deteriorates in agile scenarios where small-angle assumptions no longer hold.

    Early nonlinear approaches such as sliding-mode and backstepping controllers achieved robust flight control \cite{bouabdallah2005backstepping, madani2006backstepping, xu2006sliding}. The authors in \cite{lee2009feedback} demonstrated that exact nonlinear dynamic inversion of the translational and rotational dynamics is possible, however inherently suffers from lack of robustness. As a result, cascaded control structures with separate position and attitude controllers have become the standard \cite{voos2009nonlinear}.

    While earlier controllers were based on Euler angles, geometric attitude representations such as quaternions, are now widely adopted  \cite{tayebi2006attitude, fresk2013full}. A geometric tracking controller was proposed by \cite{lee2010geometric} to directly control the quadrotor on the manifold of the special Euclidean group SE(3). This controller ensures almost globally asymptotic stability of position, velocity, and attitude, allowing for highly-agile quadrotor maneuvers.

    Follow-up work reformulate the tracking problem as a state tracking problem and use the differential flatness property of quadrotors to derive feedforward terms
    from a reference trajectory \cite{mellinger2011minimum}.
    Differential flatness based control (DFBC) was further enhanced by aerodynamic drag models \cite{faessler2017differential} and incremental nonlinear dynamic inversion (INDI) control \cite{tal2020accurate}, enabling accurate and robust tracking of aggressive trajectories.
    However, effectively handling control input limits remains a challenge. Various control prioritization \cite{brescianini2018tilt, tal2020accurate} and control allocation \cite{faessler2016thrust, smeur2017prioritized, zaki2017trajectory} methods have been proposed to mitigate the actuator saturation effect. Nevertheless, the proportional nature of DFBC can often be over-aggressive in correcting errors, significantly degrading control performance at saturation limits \cite{sun2022comparative}. 
    Furthermore, feedback control methods inherently lack awareness of future reference states, limiting their ability to optimize control inputs over time when compared to predictive approaches.
    
    \paragraph{\textbf{Model Predictive Control}}
    Model Predictive Control (MPC) is the prevalent method that reformulates the control problem as an online optimization problem.
    However, MPC problems can quickly result in large optimization problems, that are computationally demanding and intractable for real-time applications. In particular, nonlinear-MPC (NMPC) involving a full-state nonlinear model of the quadrotor was not feasible on early-age flight control computers.
   
    Recent advances in hardware and nonlinear optimization solvers \cite{houska2011acado, andersson2019casadi} bring full-state NMPC towards real-time performance. Hence, recent work adopt NMPC with full nonlinear dynamics of the quadrotor and single rotor thrust constraints \cite{bicego2020nonlinear, foehn2021time, torrente2021data, salzmann2023real}. Two studies \cite{foehn2021time, sun2022comparative} demonstrate the ability to fully exploit the system capabilities in tracking race trajectories with up to $20~\frac{\unit{m}}{\unit{s}}$. The benchmark comparison in \cite{sun2022comparative} reveals strengths and weaknesses of this method. Using future predictions and reference points, NMPC outperforms DFBC methods by 48\% in tracking infeasible trajectories. However, computational loads are exceptionally high and NMPC suffers from numerical convergence issues. No rigorous stability proofs exist and convergence tends to fail when the current position is too far from the reference.
    The employment of real-time iteration (RTI) methods and system delays further diminish the robustness.
    
    \paragraph{\textbf{Neural Policy Control}}
    Neural policy control (NPC) is a relatively new control paradigm that emerged with the popularity of deep neural networks (DNNs). The impressive ability of DNNs in modeling highly nonlinear and complex functions allows for learning direct mappings from raw sensor observations to control outputs \cite{levine2016end}.

    Early work explored reinforcement learning (RL) methods to learn end-to-end neural control policies in simulated environments \cite{lillicrap2015continuous, levine2016end, zhang2016learning, hwangbo2017control}.
    Several works train a policy to map state observations directly to desired individual rotor thrusts \cite{zhang2016learning, hwangbo2017control, pi2020low}.
    A first real-world study by \cite{hwangbo2017control} demonstrates the ability to stabilize a quadrotor in the air, even under challenging initial conditions. 
    Due to the high sample complexity of learning-based policies, they are typically trained in simulation. However, the domain transfer of the policy from simulation to the real world is known to be hard. To overcome this challenge, training is often augmented with domain randomization techniques \cite{tobin2017domain, molchanov2019sim, loquercio2019deep, loquercio2021learning} and by abstracting control inputs to high-level commands \cite{kaufmann2018deep, kaufmann2020deep, song2021autonomous, loquercio2021learning}.

\subsection{Contribution}
We present a novel neural-augmented feedback tracking controller.
The controller (1.) inherits the stability and robustness properties of geometric feedback-based controllers and (2.) is sufficiently expressive to approximate optimal control inputs similar to NMPC, while (3.) retaining the computational efficiency and learning convenience of neural control policies. In formal terms, the controller exhibits the following properties:
\begin{enumerate}   
    \item \textit{Stability}: The neural augmented feedback controller has stability guarantees. The closed-loop system is contracting and any tracking error remains uniformly bounded. 
    \item \textit{Flexibility \& Expressivity}: The augmentation of the feedback loop by a neural operator increases the expressiveness of the controller significantly. 
    Furthermore, its parameterization is unconstrained and convex which allows for learning methods to be applied in a straight-forward manner. We demonstrate the expressivity of this model in learning optimal feedforward input considering nonlinear cost and actuator constraints.
    \item \textit{Efficiency}: The combination of a base controller and a neural augmentation results in an efficient control structure. The computational load for both, the stabilizing feedback and the neural network is low, facilitating real-time performance at high frequencies.
\end{enumerate}

We further highlight the following observations:
\begin{itemize}
    \item the controller shows state-of-the-art tracking performance with superior robustness to NMPC and DFBC
    \item we show that learning with a 'prior on stability' results in fast training convergence and does not require complicated training augmentations
    \item the controller inherits the robustness properties of the geometric controller and can be trained entirely in simulation and safely deployed in real-world settings
\end{itemize}

To our knowledge it presents the first optimal and learning-based flight controller with stability guarantees.

\section{Preliminaries}

\subsection{Nomenclature}
We represent vectors with bold lowercase letters and matrices with bold uppercase letters. All other variables are scalars. The inertial coordinate frame is defined as $\mathcal{F}_I: \{\bm{x^W}, \bm{y^W}, \bm{z^W}\}$ with $\bm{z^W}$ pointing upward opposing gravity. For the body-fixed frame $\mathcal{F}_B: \{\bm{x^B}, \bm{y^B}, \bm{z^B}\}$, the $\bm{z^B}$ vector is aligned with the collective thrust direction and $\bm{x^B}$ is pointing forward. The rotation from $\mathcal{F}_B$ to $\mathcal{F}_I$ is represented by the rotational matrix $\bm{R}(\bm{q}) = [\bm{x^B}, \bm{y^B}, \bm{z^B}] \in SO(3)$ with the unit quaternion parameterization $\bm{q}$.
Vectors with superscript $(\cdot)^{\bm{B}}$ are expressed in body-frame; those without superscript are expressed in inertial-frame.
\subsection{Quadrotor Model}

\newcommand\twolines[2]{\left[{{#1}\atop#2}\right]}

The translational dynamics of a quadrotor are given by
\begin{align}
    \bm{\dot x} &= \bm{v}, \label{eq:dynpos}\\
    \bm{\dot v} &= m^{-1} (T \bm{z^B} + \bm{f}_{ext}) - g \bm{z^W}, \label{eq:dynvel}
\end{align}

where $\bm{x}\in \mathbb{R}^3$ and $\bm{v}\in \mathbb{R}^3$ are the position and velocity in the inertia frame. $T$ is the collective thrust and $m$ total mass respectively. $g$ represents the gravitational acceleration and the external disturbance force vector $\bm{f}_{ext}$ accounts for all other unknown forces acting on the vehicle.

The rotational dynamics are given by
\begin{align}
    \bm{\dot q} &= \frac{1}{2} \bm{q} \otimes \left[ \bm{0} \atop \bm{\Omega^B} \right], \\
    \bm{\dot \Omega^B} &= \bm{J}^{-1} (\bm{\mu} + \bm{\mu}_{ext} - \bm{\Omega^B} \times \bm{J \Omega^B}),
    \label{eq:dynangvel}
\end{align}

where $\bm{\Omega^B}\in \mathbb{R}^3$ is the angular velocity in the body-fixed reference frame, and $\bm{q} = \left[q_w, q_x, q_y, q_z \right]$ is the normed quaternion attitude vector with $\otimes$ being the quaternion multiplication operator. The matrix $\bm{J}\in \mathbb{R}^{3 \times 3}$ is the vehicle's moment of inertia tensor and $\bm{\mu}\in \mathbb{R}^3$ indicates the control moment vector. The disturbance moment vector $\bm{\mu}_{ext}\in \mathbb{R}^3$ includes the model uncertainties on the body torque.

\section{Methodologies}

\subsection{Contraction Theory}
The contraction property implies ordered transient and asymptotic behavior, including existence, uniqueness and global exponential stability of an equilibrium for time-invariant systems \cite{bullo2023contraction}.

A system $\mathcal{T}$ is said to be contracting if, for any two initial conditions $x_0^1, x_0^2 \in \mathbb{R}^n$, the difference between their corresponding state sequences $(x_t^1, x_t^2)_{t \in \mathbb{N}}$ under the same input sequence $(u_t)_{t \in \mathbb{N}}$ converges exponentially. That is, for $\beta \in \mathbb{R}^+$ and $\alpha \in [0, 1)$,
\begin{align}
    |x^1_t - x^2_t| \leq \beta \alpha^t |x^1_0 - x^2_0| \quad \forall t \in \mathbb{N}.
\end{align}

This notion is similar to stability as described by Lyapunov, however, Lyapunov Theory characterizes stability w.r.t. a specific equilibrium whereas Contraction Theory does not require the explicit knowledge of an equilibrium. Both concepts imply that initial conditions are exponentially 'forgotten' and all states remain uniformly bounded.
We will use the term contraction whenever we discuss stability in this paper.

\subsection{Youla Parameterization} \label{sec:youla}

In brief, the Youla Parameterization (YP), or Q-augmentation, is a well known concept in linear control theory \cite{youla1976modern, khalil1996robust}, with extensions to nonlinear systems introduced by \cite{lu1995state}. We use this concept to explain how a combination of a \textit{free} stable system and a known stabilizing controller can be used to parameterize a more flexible and expressive set of stabilizing closed-loops \cite{wang2022learning, barbara2023learning}.

\begin{figure}[t!]
    \centering
    \includegraphics[width=0.4\textwidth]{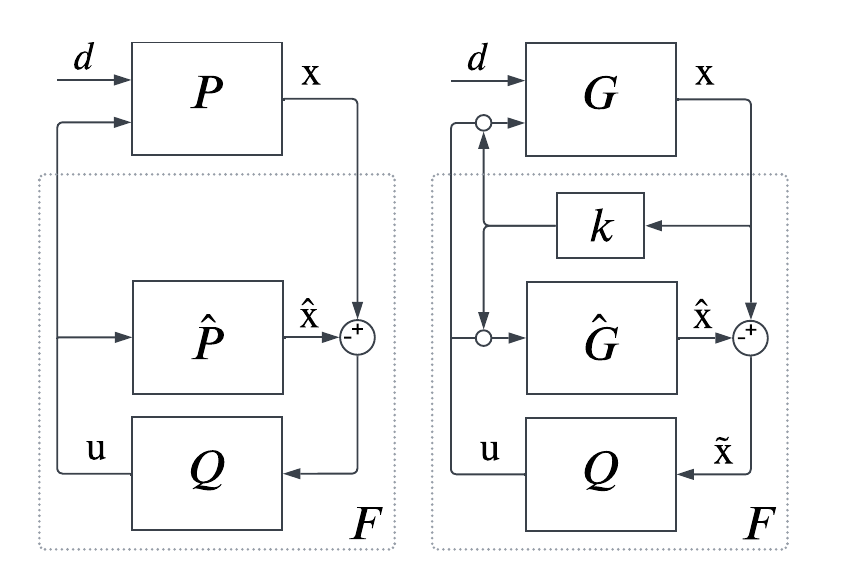}
    \caption{Block Diagram of the Youla Parameterization.}
    \label{fig:youla_feedback}
    \vspace{-3mm}
\end{figure}

For a general feedback system,
the relationship between the feedback $F$ and the closed-loop response is highly nonlinear complicating any stability analysis. An alternative representation for $F$ is presented by the Youla parameter $Q$ in combination with a nominal system $\hat{P}$ as depicted in Fig. \ref{fig:youla_feedback}.
The inclusion of a nominal model $\hat{P}$ of the system allows for the separation of state-independent terms $\Tilde{\bm{x}}$ (e.g. disturbances) from the dynamics. In this context, $Q$ and $P$ represent two systems in series.
From \cite{lu1995state}, we note that the series connection of two contracting systems $Q$ and $P$ results in an overall contracting system.
In our case, $P$ itself represents a stable system consisting of the general nonlinear system $G$ in feedback with a stabilizing controller $k$.
$Q$ is chosen as a \textit{free} system to extend the expressivity of the closed-loop. Further details including a nominal stability proof are presented in Section \ref{sec:neuralfb}.
A parameterization for $Q$ which is flexible, expressive and satisfies the contraction property is presented in the next section.

\subsection{Recurrent Equilibrium Network} \label{sec:ren}
\vspace{-1mm}
We now present a specific neural network architecture, called the recurrent equilibrium network (REN) \cite{revay2023recurrent}, a versatile neural network architecture allowing for representing all stable linear systems, all sets of contracting recurrent neural networks and all deep feedforward neural networks. Their structure is simple and resembles dynamical systems, allowing the application of control-theoretic methods. Contraction properties and Lipschitz bounds can easily be enforced by design. Hence, RENs are powerful for parameterizing \textit{flexible and expressive} sets of contracting nonlinear systems.
RENs can be succinctly described as a feedback interconnection of a linear system and a static, memoryless nonlinear operator $\sigma$:
\begin{align}
    \begin{bmatrix}
    \bm{x}_{t+1} \\ \bm{v}_t \\ \bm{y}_t
    \end{bmatrix}
    = \overbrace{\begin{bmatrix}
        \bm{A}& \bm{B}_1& \bm{B}_2 \\
        \bm{C}_1& \bm{D}_{11}& \bm{D}_{12} \\
        \bm{C}_2& \bm{D}_{21}& \bm{D}_{22}
    \end{bmatrix}}^{\bm{W}}
    \begin{bmatrix}
        \bm{x}_t \\ \bm{\omega}_t \\ \bm{u}_t
    \end{bmatrix}
    + \overbrace{\begin{bmatrix}
        \bm{b}_x \\ \bm{b}_v \\ \bm{b}_y
    \end{bmatrix}}^{\bm{b}} , \quad 
    \bm{\omega}_t = \sigma (\bm{v}_t)
\end{align}

with internal states $\bm{x}_t, \bm{x}_{t+1} \in \mathbb{R}^n$, inputs $\bm{u}_t \in \mathbb{R}^m$, outputs $\bm{y}_t \in \mathbb{R}^p$ and learnable parameters $\bm{W} \in \mathbb{R}^{(n+p+q)\times(n+m+q)}$, $\bm{b} \in \mathbb{R}^{n+p+q}$. $\bm{\omega}_t \in \mathbb{R}^q$ is the solution of an deep equilibrium network or implicit neural network:
\begin{align}
    \bm{\omega}_t = \sigma(\bm{D}_{11} \bm{w}_t + \bm{C}_1 \bm{x}_t + \bm{D}_{12} \bm{u}_t + \bm{b}_v ) \label{eq:fixedpoint}
\end{align}

and $\sigma$ is a scalar nonlinearity applied elementwise.
Eq. \ref{eq:fixedpoint} is an implicit equation for which a unique solution $\omega_t^*$ exists and can be computed efficiently. The work of \cite{revay2023recurrent} shows that a \textit{direct and unconstrained parameterization}, i.e. smooth mappings from $\mathbb{R}^n$ to the weights and biases $\theta = (\bm{W},\bm{b})$, exists that ensures the overall system is contracting with respect to its inputs $u_t$ and internal states $x_t$. Furthermore, the parameterization facilitates the direct enforcement of robustness properties in form of Lipschitz bounds, which are essential for ensuring stability amidst model discrepancies.

Using REN models, we can universally approximate all contracting and Lipschitz systems and their unconstrained parameterization allows learning methods like gradient-descent to be applied directly.

\subsection{Neural-augmented feedback control} \label{sec:neuralfb}
\vspace{-1mm}
In the following, we present the neural-augmented feedback control structure in its discrete state-space representation. 

Consider a general nonlinear system of the following form
\begin{align}
    \bm{x}_{t+1} &= G(\bm{x}_t, \bm{u}_t) + \bm{d}_t,
\end{align}
with states $\bm{x}_t \in \mathbb{R}^n$, control inputs $\bm{u}_t \in \mathbb{R}^m$ and (unknown) perturbations to the states $\bm{d}_t \in \mathbb{R}^n$. Control inputs may be modified by a known perturbation or reference input $\bm{r}_t \in \mathbb{R}^m$ so that $\bm{u}_t \mapsto \bm{u}_t + \bm{r}_t$. We assume disturbances are bounded with $|\bm{d}_t| \leq d^*$ for all $t \in \mathbb{N}$ and a $d^* \in \mathbb{R}^+$. We only discuss the full information case in this work and refer to \cite{lu1995state, barbara2023learning} for a more comprehensive discussion, including observer design. 

We assume a (robustly) stabilizing controller $\bm{u}_t = k(\bm{x}_t) + \bm{r}_t$ exists such that the closed-loop map from $\bm{d} \mapsto \bm{x}$ is contracting. In addition, the controller is augmented by a contracting neural operator $Q: \bm{\Tilde{x}} \mapsto \bm{\Tilde{u}}$. Using the YP from Sec.~\ref{sec:youla}, the neural-augmented feedback control can be stated as:
\begin{align}
    \bm{\hat{x}}_{t} &= \hat{G}(\bm{x}_{t-1}, \bm{u}_{t-1}), \\
    \bm{u}_t &= k(\bm{x}_t) + Q(\bm{\Tilde{x}}_t) + \bm{r}_t, \quad \bm{\Tilde{x}}_t = \bm{x}_t - \bm{\hat{x}}_t, 
\end{align}

with $\hat{G}$ representing the nominal dynamics model inside the controller and $\bm{\hat{x}}_{t}, \bm{x}_{t} \in \mathbb{R}$ the nominal and observed state, respectively.

It can easily be shown that for the nominal full-information (FI) case with $G = \hat{G}$ the disturbance becomes detectable in $\bm{\Tilde{x}} = \bm{d}$. Given that the perturbation $\bm{d}$ is state-independent and bounded and $Q$ represents a contracting mapping, the augmented control input $\bm{\Tilde{u}} = Q(\bm{\Tilde{x}})$ is guaranteed to be stable (contracting w.r.t. $\bm{\Tilde{x}}$) and bounded.
Note that $Q$ may involve internal states and can generate a nonlinear response $\bm{\Tilde{u}}_{t,\dots, t+n}$ based on a sequence of inputs $\bm{\Tilde{x}}_{t-n, \dots, t}$, otherwise this mapping would be trivial.

For the stability analysis of the closed-loop, the augmented control input $\bm{\Tilde{u}}$ can thus be treated as a stable and bounded exogenous input to the feedback-controlled system.
It follows by the stability criterion of the stabilizing base controller $k(\bm{x})$ that the closed-loop system will remain stable (contracting) under these conditions. Furthermore, additional state-independent and bounded terms like a reference signal $\bm{r}$ can be passed to the neural feedback as $\bm{\Tilde{u}} = Q(\bm{\Tilde{x}}, \bm{r})$ for which the same reasoning applies. This closes the nominal case.

We briefly discuss robustness w.r.t. model uncertainties at this point. For model discrepancies between the nominal and actual system $\Delta G = G - \hat{G}$, the residual state $\bm{\Tilde{x}} = \Delta G(\bm{x},\bm{u}) + \bm{d}$ includes a control-dependent term. A feedback loop exists through the neural mapping $Q$ in $\bm{u} = k(\bm{x}) + Q(\bm{\Tilde{x}})$ that could destabilize the system. Therefore, certain Lipschitz bounds of $Q$ need to be enforced to ensure passitivity of the closed-loop \cite{van2000l2}. A comprehensive discussion regarding robustness guarantees is given in \cite{wang2022youla}.

We further highlight that in practice the nominal model in the controller can be replaced by a contracting observer (ensuring that $\bm{d} \mapsto \bm{\Tilde{x}}$ is contracting) if only partial outputs of the state $\bm{x}_t$ can be observed. We refer to \cite{barbara2023learning} for the treatment of the general input-output case, including observer design.

\section{Neural Geometric Tracking Control} \label{sec:NGTC}

\begin{figure}[t]
    \centering
    \includegraphics[width=0.48\textwidth]{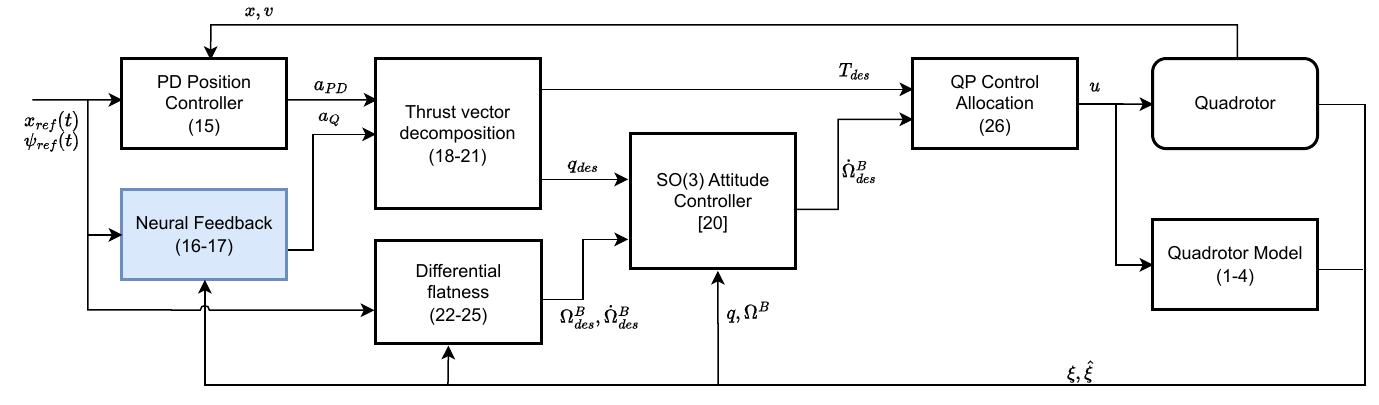}
    \caption{The control diagram of the Neural Geometric Tracking Controller.}
    \label{fig:NGTC}
    \vspace{-5mm}
\end{figure}

Building on the preceding concepts, we introduce the Neural Geometric Tracking Controller (NGTC) for agile flight.
We adopt the structure of the geometric base controller from \cite{sun2022comparative}, which is based on the differential-flatness method of \cite{mellinger2011minimum} with a tilt-prioritized attitude controller \cite{brescianini2018tilt} and quadratic-programming based control allocation \cite{johansen2013control}. The neural feedback is added to the position controller as depicted in Fig. \ref{fig:NGTC}.

\subsection{PD Position Control}

Position and velocity control is based on a proportional-derivative (PD) controller. The mathematical expression for this controller is expressed as
\begin{align}
    \bb{a}_{PD} = \bb{K}_x (\bb{x}_{ref} - \bb{x}) + \bb{K}_v (\bb{v}_{ref} - \bb{v}) + \bb{a}_{ref}
\end{align}

with $\bb{K}_x, \bb{K}_v$ as positive-definite diagonal gain matrices.

\subsection{Neural Feedback Augmentation}
\vspace{-1mm}
Following the neural-augmented feedback structure from  Sec.~\ref{sec:neuralfb}, we design the auxiliary position control input as
\vspace{-1mm}
\begin{align}
    \hat{\bm{\xi}} &= \hat{G}(\bm{\xi}_{prev}, \bm{u}_{prev}) \\
    \bm{a}_{Q} &= Q(\bm{\xi} - \hat{\bm{\xi}}, \bm{\xi}_{ref})
\end{align}

where $\hat{G}$ represents a discrete-time model of the nominal dynamics presented in (\ref{eq:dynpos}-\ref{eq:dynangvel}) with $\bm{f}_{ext} = \bm{\mu}_{ext} = \bm{0}$ and a first-order motor model. A 4th-order Runge-Kutta integrator is used for the discretization. $Q$ is a REN according to Sec.~\ref{sec:ren}. The subscript $(\cdot)_{prev}$ is used to indicate the state $\bm{\xi} = (\bm{x}, \bm{v}, \bm{q}, \bm{\Omega^B})$ and control input $\bm{u}$ from the previous timestep. $\bm{\xi}_{ref}$ is a vector of future reference states.

\subsection{Desired Attitude and Collective Thrust}
\vspace{-1mm}
From (\ref{eq:dynvel}) we obtain the desired thrust $T_{des}$ and thrust direction $\bm{z}^B_{des}$ as
\begin{align}
    T_{des} \bm{z}^B_{des} = (\bm{a}_{PD} + \bm{a}_Q - g \bm{e}_3) \: m.
\end{align}

Given the reference heading angle $\psi_{ref}$, the desired attitude can be obtained by the following equations:
\begin{align}
    \bm{y}^C_{des} &= \left[\sin(\psi_{ref}), \cos(\psi_{ref}), 0 \right]^T, \\
    \bm{x}^B_{des} &= \frac{\bm{y}^C_{des} \times \bm{z}^B_{des}}{||\bm{y}^C_{des} \times \bm{z}^B_{des}||}, \\
    \bm{R}(\bm{q}_{des}) &= \left[\bm{x}^B_{des}, \bm{z}^B_{des} \times \bm{x}^B_{des}, \bm{z}^B_{des}  \right],
\end{align}

where unit quaternion $\bm{q}_{des}$ expresses the desired attitude.
We use the tilt-prioritized attitude controller presented in \cite{brescianini2018tilt} in combination with a quadratic-programming control allocation \cite{sun2022comparative} to track the desired attitude.

\subsection{Angular Velocity Reference}
\vspace{-1mm}
We use the differential flatness property of quadrotors to derive additional feedforward reference states for the angular velocity and angular acceleration from higher derivatives of the position trajectory. The inclusion of jerk is important for tracking aggressive trajectories where attitude changes occur rapidly \cite{tal2020accurate}.

Jerk is derived by taking the derivative of (\ref{eq:dynvel}).

An expression for $\Omega^B_{x}, \Omega^B_{y}$ and $\dot T$ is obtained by projecting on to the three body-fixed axes:
\begin{equation}
    \resizebox{0.43\textwidth}{!}{
    $m \bm{R}^T \bm{j} = \dot T \bm{z^W} + T \bm{\Omega^B} \times \bm{z^W} = [T \Omega^B_{y}, - T \Omega^B_{x}, \dot T]^T,$
    }
    \label{eq:angvelref}
\end{equation}

where the subscript $(\cdot)_{ref}$ is omitted for brevity.
Any higher-order feedforward terms did not improve the tracking performance in our real-world experiments.

\section{Experiments}

In this section, experimental results for robust agile quadrotor flight are presented. We compare our approach (NGTC) to Nonlinear Model Predictive Control (NMPC) and Differential Flatness-Based Control (DFBC), two state-of-the-art control methods. 
In a set of simulated experiments, we evaluate tracking performance (\ref{sec:optimaltracking}), robustness (\ref{sec:robustness}) and  computational efficiency (\ref{sec:efficiency}).

\begin{figure}[t!]
    \centering
    \includegraphics[width=0.5\textwidth]{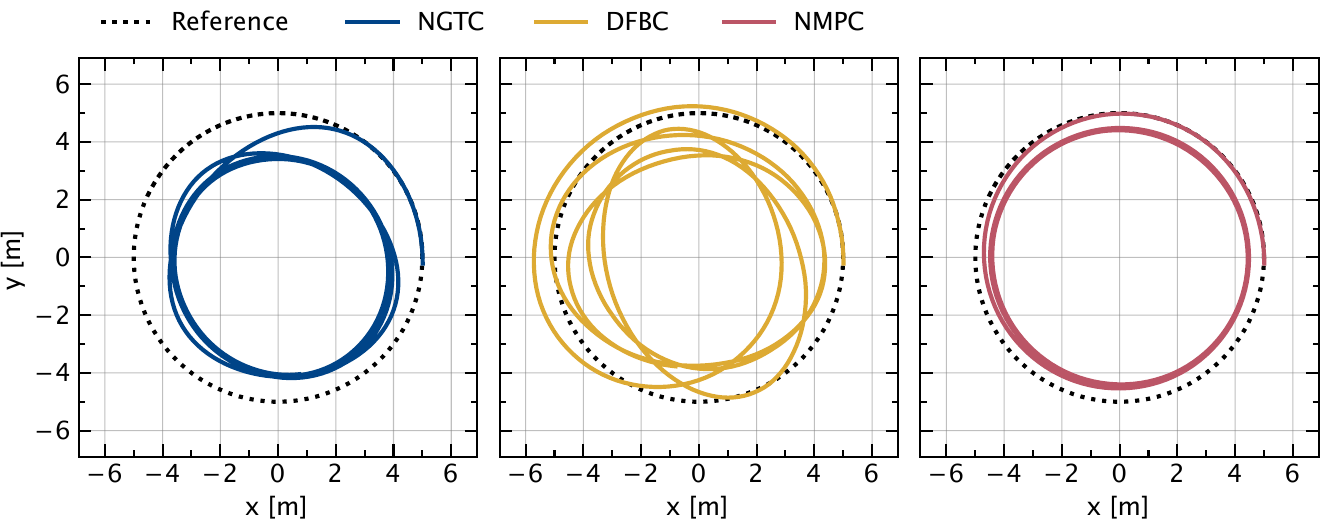}
    \caption{Position trajectories of tracking a dynamically infeasible circular trajectory ($V_{max} = 15 m/s, a_{max} = 45 m/s^2$). Depicted are 2D trajectories from top view.
    }
    \label{fig:circular}
    \vspace{-3mm}
\end{figure}

\subsection{Implementation Details}
\newcommand{\exnum}[2]{{#1}\mathrm{e}{#2}}

\begin{table}[t]
    \centering
    \captionsetup{font=footnotesize}
    \caption{Quadrotor model parameters, costs and gains.}
    \label{tab:quadparams}
    \scalebox{0.94}{
   \begin{tabular}{ll|ll}
    \toprule
       \multicolumn{2}{c|}{NMPC} & \multicolumn{2}{c}{DFBC, NGTC}
         \\ [0.5ex] \midrule
         $\bm{Q}_x$&(200, 200, 500)&$\bm{K}_x$&(18, 18, 18)\\
         $\bm{Q}_v$&(1, 1, 1)&$\bm{K}_v$&(8, 8, 8)\\
         $\bm{Q}_q$&(5, 5, 200)&$(k_{q,xy}, k_{q,z})$&(150, 3)\\
         $\bm{Q}_\Omega$&(1, 1, 1)&$\bm{K}_\Omega$&(20, 20, 8)\\
         $\bm{Q}_u$&(6, 6, 6, 6)&$\bm{W}_c$&$(1\mathrm{e}-3, 10, 10, 0.1)$\\
         ($dt$, $N$)& (50 ms, 20)&$REN(n, m, q, p)$&(32, 96, 256, 3)
         \\ \bottomrule \toprule
         \multicolumn{4}{c}{
         \begin{tabular}{c|c|c|c|c}
             $m$ & $l$ & $\beta$ & $\bm{J}$ $[gm^2]$ & $(u_{min}, u_{max})$\\ \midrule
            0.72 [kg]&0.14 [m]&56 [°]&diag(2.5, 2.1, 4.3)&(0, 8.5) [N]
         \end{tabular}
         }
         \\ \bottomrule
    \end{tabular}
    }
\end{table}

\paragraph{Simulation}
We use a 4th-order Runge-Kutta integrator running at 100 Hz to integrate the quadrotor dynamics defined in equations (\ref{eq:dynpos})-(\ref{eq:dynangvel}). The inertial and geometric parameters of the quadrotor model are listed in Table \ref{tab:quadparams}. Motor dynamics are simulated by passing the commanded rotor thrust commands through a first-order low-pass filter with a 30ms time constant $\tau_{mot}$. States are directly fed into the controller without simulating any estimation errors. The fidelity of the simulation is further increased by an aerodynamic drag model (from \cite{sun2022comparative}).
\paragraph{Controller}
Both geometric controllers NGTC and DFBC follow the implementation presented in Sec. \ref{sec:NGTC} with and without neural feedback augmentation, respectively. The neural feedback model is represented by an acyclic REN \cite{revay2023recurrent} where $\bm{D}_{11}$ is strictly lower triangular to ensure fixed execution time of the implicit equation. Model parameter dimensions and gains are listed in Table \ref{tab:quadparams}. For NMPC, we adopt the implementation of \cite{sun2022comparative} for which the cost are specified in Table \ref{tab:quadparams}. None of the controllers include an aerodynamic drag model.
\paragraph{Real-world}
All controllers are implemented in the Agilicious \cite{foehn2022agilicious} software framework and executed on a Jetson Orin Nano compute platform on the drone. Collective thrust and attitude rate commands are send to a lower-level flight controller that handles communication with the ESCs.

\subsection{Training}
We train the NGTC using Analytic Policy Gradient \cite{wiedemann2023training} in a PyTorch environment. The inherent stability of our approach allows us to simply forward simulate the controlled system, collect trajectories and use backpropagation through time (BPTT) to update model parameters. The optimal control cost of NMPC (\ref{tab:quadparams}) is selected as training loss. We train the model on a diverse set of 100k Lissajous trajectories defined by:
\begin{align}
\begin{split}
    \bm{x}_{ref}(t) = [ A_x \sin(\omega_x t + \omega_{x,0}), A_y \sin(\omega_y + \omega_{y,0}), \\ A_z \sin(\omega_z t + \omega_{z,0}) ]^T,
\end{split}
\end{align}
with $A_x, A_y \in [0,20]$, $A_z \in [0,3]$ and $\omega_x, \omega_y, \omega_z \in [0,5]$, $\omega_{x,0},\omega_{y,0},\omega_{z,0} \in [0,2\pi]$. The yaw reference $\psi_{ref}$ is chosen to always point along the direction of the velocity vector. After sampling, the dataset is filtered for actuator feasibility and trajectories exceeding motor thrust limits by more than $10\%$ are excluded. For both, the simulator and controller, only nominal model parameters are used (see Tab.~\ref{tab:quadparams}) during training. For each run, the disturbance terms $\bm{f}_{ext}$ and $\bm{\mu}_{ext}$ are modeled as random constant force ($<$20N) and torque vectors ($<$0.1Nm) with additional 20\% Gaussian noise variations. No aerodynamic drag is simulated during training time to ensure a fair comparison.

\subsection{Tracking performance} \label{sec:optimaltracking}
We present quantitative results on tracking a selection of feasible and infeasible trajectories in Table \ref{tab:accuracy}. In addition, a qualitative comparison is done to highlight the improvements on tracking infeasible trajectories.

\subsubsection{Tracking Accuracy}
All three methods NMPC, DFBC and NGTC perform equally well under nominal conditions with perfect model knowledge and state estimates which has also been highlighted in \cite{sun2022comparative} (three top rows in Tab.~\ref{tab:accuracy}). For DFBC, tracking accuracy drops significantly when trajectories become dynamically infeasible. NGTC improves tracking accuracy by at least 40\% under these conditions, however NMPC still outperforms both methods on most trajectories (three bottom rows in Tab.~\ref{tab:accuracy}).

\begin{table}[t]
    \centering
    \captionsetup{font=footnotesize}
    \caption{Position 
    RMSE for tracking dynamically feasible and infeasible~{($^*$)} trajectories. 3D trajectory paths are shown in \cite{sun2022comparative}.}
    \label{tab:accuracy}
    \scalebox{0.95}{
    \begin{tabular}{l|>{\centering\arraybackslash}p{1.6cm} >{\centering\arraybackslash}p{1.6cm} >{\centering\arraybackslash}p{1.6cm}}
    \toprule
    &  \multicolumn{3}{c}{Position RMSE [m]} \\
    & DFBC & NMPC & NGTC (ours) 
        \\ [0.5ex] \midrule
     Hor. Loop & 0.25 & 0.23 & $\bm{0.20}$ \\
     Ver. Loop & 0.20 & 0.18 & $\bm{0.17}$ \\
     Lemniscate & $\bm{0.14}$ & 0.22 & 0.15 \\
     \midrule
     Hor. Loop$^*$ & 2.39 & 1.77 & $\bm{1.42}$ \\
     Ver. Loop$^*$ & 5.47 & $\bm{1.06}$ & 1.19 \\
     Lemniscate$^*$ & 2.04 & $\bm{0.84}$ & 1.13 \\
      \bottomrule
    \end{tabular}
    }
\end{table}

\subsubsection{Tracking infeasible Trajectories}

We qualitatively compare all three controllers on tracking an infeasible circular trajectory in Fig. \ref{fig:circular}. Tracking this reference requires a collective thrust input that exceeds the actuator bounds. Hence the quadrotor can not follow the reference directly and all three methods shortcut to fly inside the reference trajectory. Both NGTC and NMPC follow a smaller inner circle trajectory while DFBC results in a chaotic trajectory. NGTC learns feedforward inputs that allow it to track a feasible inner circle. The smaller radius can be attributed to the disturbance-prone training setting.
Our results are consistent with the real-world experiments performed in \cite{sun2022comparative}.

\subsection{Stability \& Robustness} \label{sec:robustness}
To further examine the robustness of all three methods, we evaluate their tracking performance in presence of external disturbances and model mismatch.

\subsubsection{Model Discrepancy}
Table \ref{tab:robustness} shows a comparison of the tracking accuracy for different model discrepancies. Across all perturbations, NGTC demonstrates increased robustness with respect to DFBC and NMPC.

\begingroup
\setlength{\tabcolsep}{2pt}
\begin{table}[t]
    \centering
    \captionsetup{font=footnotesize}
    \caption{Comparison of tracking errors with model mismatches affecting translational dynamics. The values represent the mean and standard deviation (crashes excluded) across tracking 30 feasible Lissajous trajectories.}
    \label{tab:robustness}
    \scalebox{0.9}{
    \begin{tabular}{l|cr@{\hskip 10pt}cr@{\hskip 10pt}cr}
    \toprule
        &  \multicolumn{6}{c}{Position RMSE [m] \quad (crash rate)} \\
        & \multicolumn{2}{c}{DFBC} & \multicolumn{2}{c}{NMPC} & \multicolumn{2}{c}{NGTC (ours)} 
         \\ \midrule
      $+ 50\%$ Drag & 0.74 $\pm 0.11$ & $(7\%)$ & 0.73 $\pm$ 0.15 & $(0\%)$ & $\bm{0.67}$ $\pm 0.16$ & $(0\%)$ \\
      $+ 30\%\ \tau_{mot}$ & $0.49 \pm 0.11$ & $(0\%)$ & $\bm{0.42} \pm 0.09$ & $(0\%)$ & $0.44 \pm 0.12$ & $(0\%)$ \\
      $- 30\%$ Mass & $0.71 \pm 0.32$ & $(3\%)$ & $0.73 \pm 0.19$ & $(3\%)$ & $\bm{0.59} \pm 0.12$ & $(\bm{0\%})$ \\
      $+ 30\%$ Mass & $0.89 \pm 0.69$ & $(7\%)$ & $0.78 \pm 0.20$ & $(10\%)$ & $\bm{0.62} \pm 0.14$ & $(\bm{0\%})$ \\
      10N ext. Force & $0.33 \pm 0.12$ & $(0\%)$ & $0.29 \pm 0.13$ & $(0\%)$ & $\bm{0.22} \pm 0.08$ & $(0\%)$ \\
      15N ext. Force & $0.65 \pm 0.24$ & $(7\%)$ & $0.71 \pm 0.18$ & $(27\%)$ & $\bm{0.34} \pm 0.09$ & $\bm{(3\%)}$ \\ \bottomrule
    \end{tabular}
    }
    \vspace{-3mm}
\end{table}
\endgroup
\subsubsection{External Disturbances}

\begin{figure}[b!]
    \centering
    \vspace{-5mm}
    \includegraphics[width=0.35\textwidth]{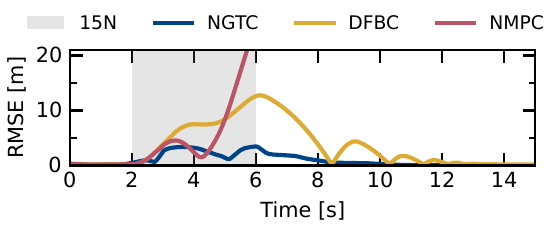}
    \vspace{-2mm}
    \caption{Position tracking error for a circular trajectory ($V_{max} = 15 m/s$, $a_{max} = 40 m/s^2$). The gray area indicates a period where a lateral external disturbance force $\bm{f}_{ext} = 15N$ is acting on the drone.}
    \label{fig:disturbance}
\end{figure}

Fig. \ref{fig:disturbance} shows an example of the position tracking error for all three controllers when experiencing a 15 N external disturbance force. Both NMPC and DFBC deviate significantly from the reference during the disturbance period. NMPC fails to converge and becomes instable while DFBC is able to recover after the disturbance. NGTC remains stable with much lower tracking error indicating improved disturbance rejection capabilities.

\begin{figure}[t!]
    \centering
    \includegraphics[width=0.49\textwidth]{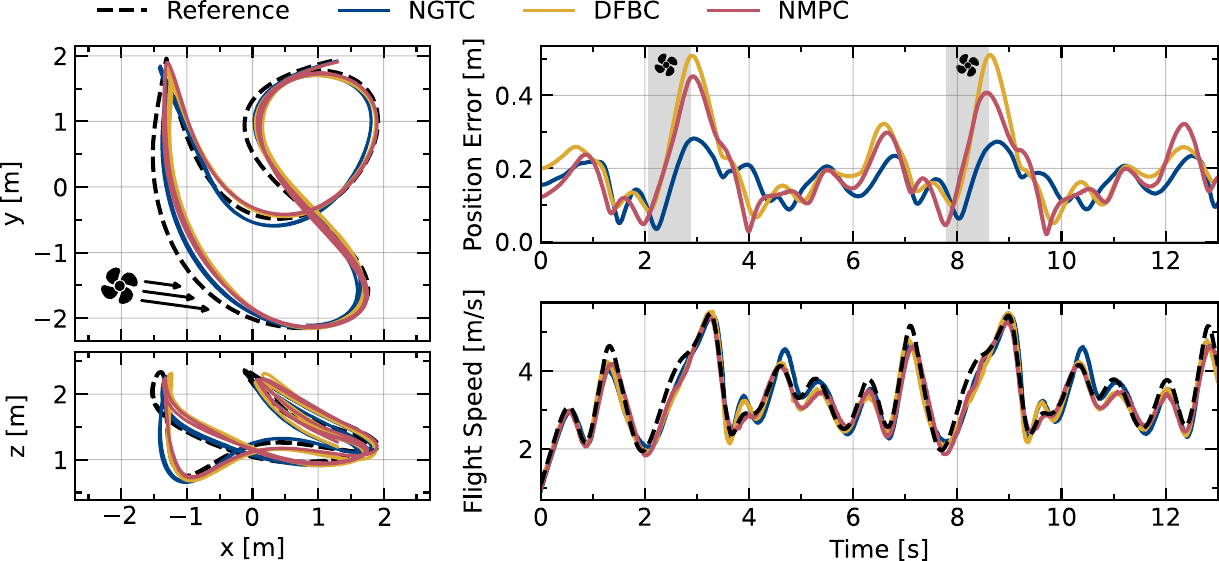}
    \caption{Real-world tracking of a dynamic racing trajectory. Left: top and side views of the reference trajectory and tracking results for the three controllers; Top-right: Norm of position tracking error over time; Bottom-right: Norm of velocity in the reference case and for the three controllers.}
    \label{fig:tracking}
    \vspace{-3mm}
\end{figure}

\ifx\shortVersion
The robustness is further examined with respect to different disturbance magnitudes in Fig.~\ref{fig:disturbance_mag}. For small disturbances ($<10 N $), all three controllers track the reference accurately. As disturbances increase, the RMSE for DFBC grows proportionally. NMPC demonstrates better tracking accuracy for higher external forces but suddenly becomes instable when deviations exceed a certain threshold. NGTC demonstrates robust and accurate tracking, even under strong disturbances.
\fi

\subsection{Efficiency} \label{sec:efficiency}

The computational efficiency is evaluated by comparing the respective processing times to generate the control command. Therefore, we compare C++ implementations of both NMPC and DFBC from the \textit{Agilicious} framework \cite{foehn2022agilicious}. For NGTC, we add the fixed evaluation time of the REN to the processing time of DFBC. All experiments are done on a 1.8 GHz Intel Core i7 processor using a single CPU core.

DFBC shows execution times that are generally faster than 0.025 ms. For NGTC, the additional forward pass through the REN accounts for 0.576 ms total execution time, which is still significantly faster compared to NMPC which requires around 3 ms.

\subsection{Learning with a prior on stability}
We evaluate the training stability of APG (Analytic Policy Gradient) by comparing our method to conventional neural policies. For an end-to-end neural control policy, the training without curriculum learning is instable. Training with curriculum learning is slower and requires more iterations. In contrast, the NGTC including a contracting REN demonstrates fast and stable convergence.

\subsection{Real-world Tracking with Disturbance}
The tracking performance on a real quadrotor (Fig.~\ref{fig:agileflight}) is evaluated by tracking an agile racing trajectory while subject to a strong external wind disturbance. The experiment is performed in an instrumented flight arena where accurate position measurements are available. The respective position trajectories and the corresponding tracking error for NGTC, DFBC and NMPC are shown in Fig.~\ref{fig:tracking}. For both NMPC and DFBC the wind disturbance results in a significant deviation from the reference while NGTC is more effective in counteracting the disturbance and achieves a lower tracking error. These results highlight the effectiveness of NGTC in real-world conditions, reinforcing its potential for robust performance in dynamic, high-speed flight scenarios.

\section{Conclusion}
We presented a novel geometric tracking controller with neural feedback augmentation for agile flight, achieving state-of-the-art tracking accuracy, particularly when subject to external disturbances or on dynamically infeasible trajectories, compared to NMPC and DFBC. Our approach enhanced robustness, improving disturbance rejection and tracking accuracy. The inherent stability of the control design facilitates real-world transfer and ensures rapid and reliable convergence during training. Furthermore, the controller's computational efficiency, with significantly shorter execution times than NMPC, makes it well-suited for real-time deployment in high-speed, safety-critical applications.

\printbibliography{}
\typeout{get arXiv to do 4 passes: Label(s) may have changed. Rerun}
\end{document}